\documentclass[twoside]{article}

\usepackage[accepted]{aistats2022}

\usepackage{algorithm}
\usepackage{algorithmic}
\usepackage{graphicx}
\usepackage{amsmath}
\usepackage{amssymb}

\newcommand{\xvec}{{\bf x}}

\usepackage{subfigure}

\usepackage[round]{natbib}

\begin{document}

%

%

\twocolumn[

\aistatstitle{Co-Regularized Adversarial Learning for Multi-Domain Text Classification}

\aistatsauthor{ Yuan Wu \And Diana Inkpen \And  Ahmed El-Roby }

\aistatsaddress{ Carleton University \And  University of Ottawa \And Carleton University } ]

\begin{abstract}
  Multi-domain text classification (MDTC) aims to leverage all available resources from multiple domains to learn a predictive model that can generalize well on these domains. Recently, many MDTC methods adopt adversarial learning, shared-private paradigm, and entropy minimization to yield state-of-the-art results. However, these approaches face three issues: (1) Minimizing domain divergence can not fully guarantee the success of domain alignment; (2) Aligning marginal feature distributions can not fully guarantee the discriminability of the learned features; (3) Standard entropy minimization may make the predictions on unlabeled data over-confident, deteriorating the discriminability of the learned features. In order to address the above issues, we propose a co-regularized adversarial learning (CRAL) mechanism for MDTC. This approach constructs two diverse shared latent spaces, performs domain alignment in each of them, and punishes the disagreements of these two alignments with respect to the predictions on unlabeled data. Moreover, virtual adversarial training (VAT) with entropy minimization is incorporated to impose consistency regularization to the CRAL method. Experiments show that our model outperforms state-of-the-art methods on two MDTC benchmarks. 
\end{abstract}

\section{Introduction}

Text classification is a fundamental task in natural language processing (NLP) \citep{young2018recent} and has attracted constant attention in recent years, it has been widely used to reshape business via understanding customers' emotional tendency \citep{smith2017two} and conduct spam detection \citep{ngai2011application}. With the advent of deep learning, text classification obtains impressive achievements in various applications \citep{kowsari2019text,wu2021mixup}. However, these achievements always rely on large amounts of labeled training data. In many real-world scenarios, abundant labeled training data are not commonly available and data labeling is always expensive and time-consuming. Therefore, it is of great significance to investigate how to improve the classification accuracy of the target domain by leveraging available resources from related domains. 

Currently, there are three categories to address the above problem. The first one is the domain-agnostic method, which combines all available labeled data from existing domains as the training set, ignoring domain differences. Unfortunately, text classification is a highly domain-dependent task where the same word in different domains may express different sentiments. For example, the word \textit{easy} frequently indicates positive sentiment in an electronic device review (e.g. the camera is easy to use), while expressing negative remark in a movie review (e.g. the ending of this movie is easy to guess). Thus, this method fails to produce a satisfactory result \citep{chen2018multinomial}. The second one is to fine-tune large pre-trained language models \citep{zheng2018same}, such as BERT \citep{devlin2019bert} and OpenAI GPT \citep{radford2018improving}, on the target domain. The large language models can effectively express context information of a word compared to general word embedding and yield remarkable performance in various NLP tasks \citep{devlin2019bert}. The third one is multi-domain text classification (MDTC) \citep{li2008multi}, which tackles the scenario where texts come from multiple domains, each with limited amounts of labeled data and large amounts of unlabeled data. Many recent MDTC methods \citep{liu2017adversarial,chen2018multinomial,zheng2018same,wu2020dual,wu2021conditional,wu2021mixup} adopt the adversarial learning, shared-private paradigm, and entropy minimization to yield state-of-the-art results. Adversarial learning performs feature alignment through reducing domain divergence to learn domain-invariant features, these features are supposed to be both transferable and discriminative \citep{ganin2016domain,wu2021towards}. Shared-private paradigm consists of two types of feature extractors: the shared feature extractor constructs the shared latent space to learn domain-invariant features, and a set of domain-specific feature extractors, one per domain, each of which captures domain-specific knowledge \citep{liu2017adversarial}. Entropy minimization is often employed to regularize the model output and minimize the uncertainty of predictions on unlabeled data \citep{long2018conditional,wu2021conditional}. However, these MDTC methods still face three major issues: First, minimizing domain divergence can not fully guarantee the success of domain alignment. Second, aligning marginal feature distributions may match samples with different classes, leading to corrupted classification accuracy. Third, entropy minimization may make the predictions on unlabeled data over-confident, deteriorating the discriminability of the learned feature representations.

In this paper, we propose a co-regularized adversarial learning (CRAL) mechanism to alleviate the aforementioned issues. The proposed CRAL method constructs two diverse shared latent spaces and aligns class-conditional feature distributions in each of them to improve the system performance. These two alignments are enforced to agree with each other with respect to the predictions on unlabeled data, which helps in shrinking the search space of possible alignments while preserving the correct set of alignments. Moreover, virtual adversarial training (VAT) \citep{miyato2018virtual} with entropy minimization is also incorporated to impose the consistency regularization to our model, ruling out redundant hypothesis classes that abruptly change predictions in the vicinity of the training data \citep{chapelle2005semi}. Experimental results show that our proposed approach can outperform the state-of-the-art MDTC methods on two benchmarks. Further experiments on the multi-source unsupervised domain adaptation (MS-UDA) scenario where we train the model on multiple domains and evaluate the model on an unseen domain demonstrate that CRAL has a good ability to generalize the learned knowledge to unseen domains. The contributions of our work are listed as follows:

\begin{itemize}
    \item We propose a co-regularized adversarial learning (CRAL) method for multi-domain text classification which constructs two diverse shared latent spaces, performs adversarial alignment in each of them, penalizes the disagreements of their predictions on unlabeled data, and incorporates virtual adversarial training with entropy minimization to improve the system performance.
    
    \item We demonstrate the effectiveness of the CRAL method on two MDTC benchmarks. The experimental results show that our method outperforms the state-of-the-art methods on both datasets. Further experiments on multi-source unsupervised domain adaptation reveal the generalization ability of our model.
    
    \item We also conduct ablation studies and parameter sensitivity analyses to explore the contributions of different components of the CRAL method and how each hyperparameter influences the performance of our method.
\end{itemize}

\section{Related Work}

Multi-Domain Text Classification aims to leverage available resources from multiple domains to improve the classification accuracy over all domains \citep{li2008multi}. More recently, deep neural networks have significantly advanced the performance of MDTC models. The multi-task convolutional neural network (MT-CNN) uses a convolutional layer to enable better word embedding \citep{collobert2008unified}. The collaborative multi-domain sentiment classification (CMSC) approach constructs domain-specific predictive modules to enhance the predictions of the shared classifier \citep{wu2015collaborative}. The multi-task deep neural network (MT-DNN) establishes a low-dimensional latent space to generate semantic vector representations for the downstream classification. \citep{liu2015representation}. 

Adversarial Training is first proposed in the generative adversarial network (GAN) for image generation \citep{goodfellow2014generative}, it adversarially trains a discriminator against a generator: the discriminator contrives to distinguish real images from generated images, while the generator struggles to fool the discriminator. Then adversarial training is extended to learn domain-invariant features in domain adaptation \citep{ganin2016domain,zhao2017multiple,wu2020dual1}. \cite{bousmalis2016domain} proposes a shared-private paradigm to reveal that domain-specific knowledge can capture the unique characteristics of its own domain and complements domain-invariant features to enhance their discriminability. Entropy and confidence are reasonable selection criteria for controlling uncertainty of predictions on unlabeled data \citep{grandvalet2005semi}, minimizing the entropy of predictions on unlabeled data can impose high priority to "easy-to-transfer" instances, facilitating the domain alignment \citep{long2018conditional}. 

Many state-of-the-art MDTC approaches adopt adversarial learning, shared-private paradigm, and entropy minimization. The adversarial multi-task learning for text classification (ASP-MTL) utilizes orthogonality constraints to enhance the separations of domain-invariant features and domain-specific features such that different aspects of the inputs can be encoded \citep{liu2017adversarial}. The multinomial adversarial network (MAN) derives the generalization bounds for adversarial multi-domain text classification with respect to the least square loss and the negative log-likelihood loss \citep{chen2018multinomial}. The dynamic attentive sentence encoding (DA-MTL) method utilizes the attention mechanism and introduces a task-dependent query vector to learn domain-specific features to enhance the discriminability of the shared features \citep{zheng2018same}. The dual adversarial co-learning (DACL) approach combines the discriminator-based adversarial learning and the classifier-based adversarial learning to enhance the discriminability of the domain-invariant features \citep{wu2020dual}. The global and local shared representation based generic dual-channel multi-task learning (GLR-MTL) method deploys adversarial training and mixture-of-experts on two separate channels to capture global-shared, local-shared, and private features simultaneously \citep{su2020multi}.  The conditional adversarial networks (CANs) conduct alignment on joint distributions of domain-invariant features and predictions to enhance the discriminability of the learned features, and use entropy conditioning to avoid the risk of conditioning on predictions with low certainty \citep{wu2021conditional}. 

In contrast with the prior MDTC approaches, our proposed CRAL method penalizes the disagreement of predictions on unlabeled data induced from two independent adversarial training streams to boost the discriminability of the learned features. Moreover, it introduces the virtual adversarial training with entropy minimization to drive the decision boundary away from the high-density regions of distributions, imposing consistent prediction constraints to the model.

\section{Method}

The MDTC setting is formulated as follows: Given $M$ different domains $\{D_i\}_{i=1}^M$, $D_i$ consists of two parts: a limited amount of labeled instances $\mathbb{L}_i=\{(\xvec_j,y_j)\}_{j=1}^{l_i}$, and a set of unlabeled instances $\mathbb{U}_i=\{\xvec_j\}_{j=1}^{u_i}$, where $l_i$ is the number of the labeled samples and $u_i$ is the number of the unlabeled samples. The main objective of MDTC is to improve the average classification accuracy across the $M$ domains by leveraging all available resources.


\begin{figure*}
    \centering
    \includegraphics[width=0.7\columnwidth]{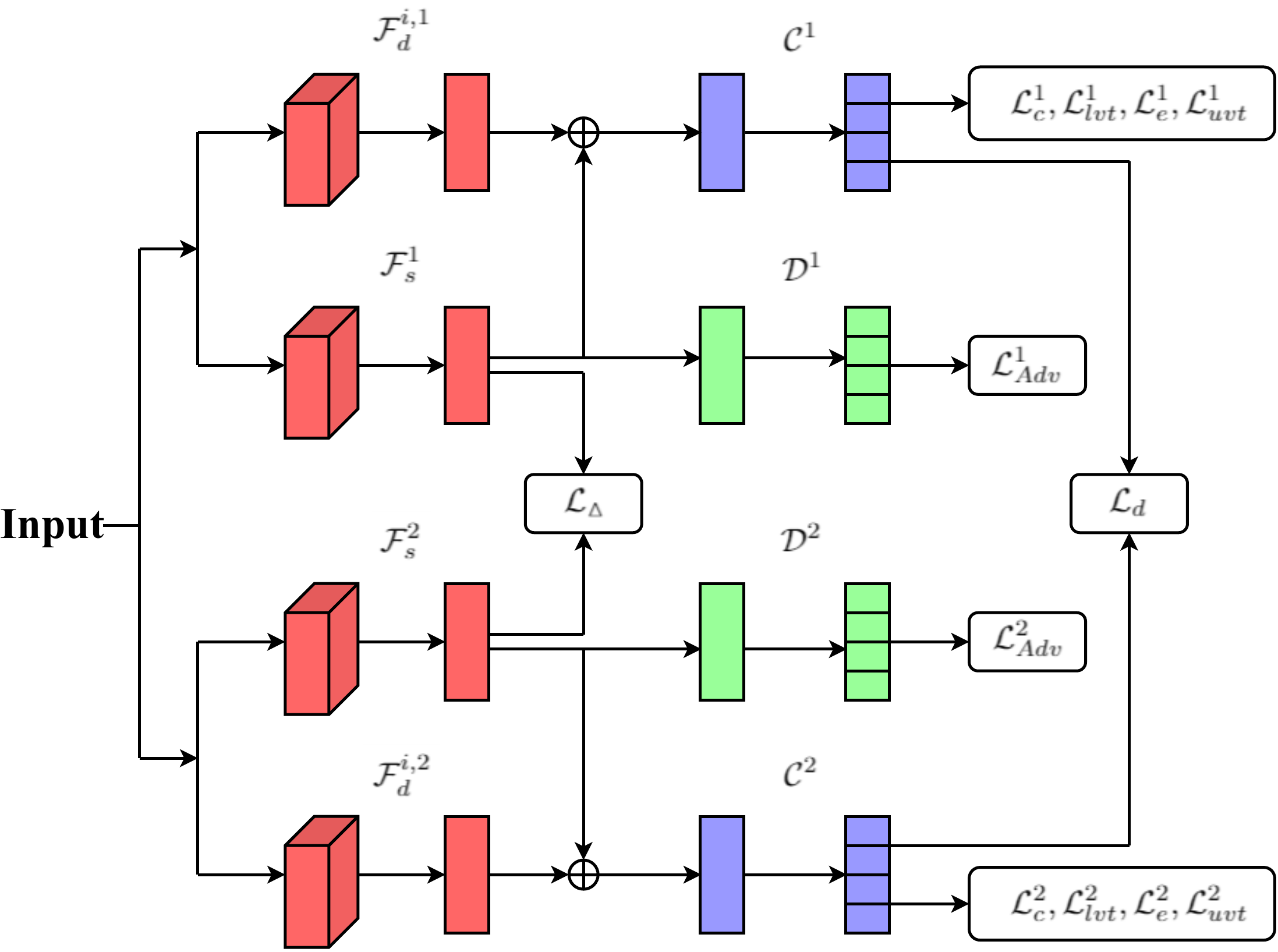}
    \caption{The Architecture of the CRAL framework. The shared feature extractors $\{\mathcal{F}_s^b\}_{b=1}^2$ are used to capture domain-invariant features, the domain-specific feature extractors $\{\mathcal{F}_d^{i,b}\}_{b=1}^2$ aim to learn domain-dependent features, the domain discriminators $\{\mathcal{D}^b\}_{b=1}^2$ are used to distinguish features across domains, and the classifiers $\{\mathcal{C}^b\}_{b=1}^2$ are used to conduct text classification. $\{\mathcal{L}_c^b\}_{b=1}^2$ are classification losses, $\{\mathcal{L}_{Adv}^b\}_{b=1}^2$ are adversarial losses, $\{\mathcal{L}_e^b\}_{b=1}^2$ are entropies of unlabeled data, $\{\mathcal{L}_{uvt}^b\}_{b=1}^2$ are VAT losses of unlabeled data, $\{\mathcal{L}_{lvt}^b\}_{b=1}^2$ are VAT losses of labeled data, $\mathcal{L}_\delta$ measures the diversity between the two shared feature extractors, and $\mathcal{L}_d$ measures the disagreement of predictions of the two classifiers on unlabeled data.} 
    \label{Fig1}
\end{figure*}


\subsection{Model Architecture}

As illustrated in Fig.\ref{Fig1}, the proposed CRAL model consists of two branches, each branch is composed of four components: a shared feature extractor $\mathcal{F}_s^b$, a set of domain-specific feature extractors $\{\mathcal{F}_d^{i,b}\}_{i=1}^M$, a domain discriminator $\mathcal{D}^b$, and a classifier $\mathcal{C}^b$. $b$ indicates the branch index (e.g $b\in\{1,2\}$). For simplicity, we ignore the branch index of each component, the following discussion holds for both branches. The shared feature extractor aims to learn domain-invariant features that can generalize across domains. The domain-specific feature extractor captures domain-specific features that are beneficial within their own domain. These feature extractors can adopt the architecture of convolutional neural networks (CNNs), recurrent neural networks (RNNs), transformer, or multi-layer perceptrons (MLPs), depending on the practical task. The feature extractors generate feature representations with the fixed length. The discriminator takes a shared feature vector as its input, while the classifier uses the concatenation of a shared feature and a domain-specific feature.

\subsection{Multinomial Adversarial Learning}

Multinomial adversarial learning minimizes the f-divergence \citep{nowozin2016f} among multiple domains to conduct alignment and has been widely applied in MDTC \citep{wu2020dual,wu2021mixup,wu2021conditional}. In multinomial adversarial learning, the domain discriminator and the shared feature extractor are trained to compete against each other: the domain discriminator tries to distinguish features among different domains, while the shared feature extractor aims to confuse the domain discriminator. When these two components reach equilibrium, the learned features can be regarded as domain-invariant. In the CRAL method, multinomial adversarial learning is performed independently in each of the two branches. $\mathcal{D}$ is a $M$-class classifier, outputting the probabilities of an instance coming from each domain. $\mathcal{C}$ is a binary classifier, predicting sentiment probabilities. $\mathcal{F}_s$ and $\{\mathcal{F}_d^i\}_{i=1}^M$ are expected to complement each other and maximally capture useful information across domains. The multinomial adversarial learning is encoded as follows:

\begin{align}
    \mathcal{L}_c^b = -\sum_{i=1}^M \mathbb{E}_{(\xvec_i,y_i)\sim\mathbb{L}_i} [y_i^{\top}\log(\mathcal{G}^{i,b}(\xvec_i))]
\end{align}

\begin{align}
    \mathcal{L}_{Adv}^b = -\sum_{i=1}^M \mathbb{E}_{\xvec_i\sim\mathbb{L}_i\cup\mathbb{U}_i}[d_i^{\top}\log(\mathcal{D}^b(\mathcal{F}_s^b(\xvec_i)))]
\end{align}

\noindent where $\mathcal{L}_c^b$ is the cross-entropy loss for the labeled data, $\mathcal{L}_{Adv}^b$ is the adversarial loss, $d_i$ is the domain index of an instance $\xvec_i$, and  $\mathcal{G}^{i,b}(\xvec_i)=\mathcal{C}^b([\mathcal{F}_s^b(\xvec_i),\mathcal{F}_d^{i,b}(\xvec_i)])$ where $[\cdot,\cdot]$ represents the concatenation of two vectors.

\subsection{Co-Regularized Adversarial Learning}

Before presenting the co-regularized adversarial learning mechanism, we first highlight the limitations of existing MDTC methods. The idea of aligning multiple feature distributions for MDTC can be motivated by the theory \citep{chen2018multinomial}: First, consider the shared features $\mathbf{f}$ extracted from each domain:

\begin{equation}
    \begin{aligned}
     P_i(\mathbf{f})\triangleq P(\mathbf{f}=\mathcal{F}_s(\xvec)|\xvec\in D_i)
    \end{aligned}
\end{equation}

\noindent where $P(\cdot)$ represents the probability, so we have $P_1$, $P_2$, ..., $P_M$ as the $M$ shared feature distributions, and set $\bar{P} = \frac{\sum_{i=1}^M P_i}{M}$ as the centroid of the $M$ feature distributions. The domain discriminator $D$ can be trained to its optimality $\mathcal{D}^*$ if and only if $P_1=P_2=...=P_M=\bar{P}$. Therefore, the objective of MDTC can be regarded as aligning $M$ feature distributions $\{P_i\}_{i=1}^M$ to $\bar{P}$ which is equivalent to minimizing the distribution distances between each of the feature distributions and their centroid. Let $\mathcal{H}$ be the hypothesis space, for any $h\in\mathcal{H}$, that maps inputs from the input space $X$ to the label space $Y$, suppose the mapping function $h=c\circ f$ can be decomposed as the composite of a feature mapping function $f:X \rightarrow \mathbb{R}^m$ and a classifier $c: \mathbb{R}^m \rightarrow Y$. Then let $h'$ be the labeling function, which can be decomposed as $h'=c'\circ f'$. Motivated by the domain adaptation theory \citep{ben2010theory}, the distance between $P_i$ and $\bar{P}$ can be defined as:

\begin{equation}
    \begin{aligned}
     d_{\mathcal{H}}^i(P_i, \bar{P})&=2\mathrm{sup}_{h,h' \in\mathcal{H}}|\mathbb{E}_{\mathbf{f}\sim P_i}[c(\mathbf{f})\neq c'(\mathbf{f})] \\
     &-\mathbb{E}_{\mathbf{f}\sim \bar{P}}[c(\mathbf{f})\neq c'(\mathbf{f})]|
    \end{aligned}
\end{equation}

\noindent The MDTC task can be treated as the collection of $M$ domain adaptation tasks between $\{P_i\}_{i=1}^M$ and $\bar{P}$, we thus obtain the total distribution distance of MDTC as:

\begin{align}
    d_{\mathcal{H}} = \sum_{i=1}^M d_{\mathcal{H}}^i(P_i, \bar{P})
\end{align}

\begin{figure*}
    \centering
    \includegraphics[width=1.0\columnwidth]{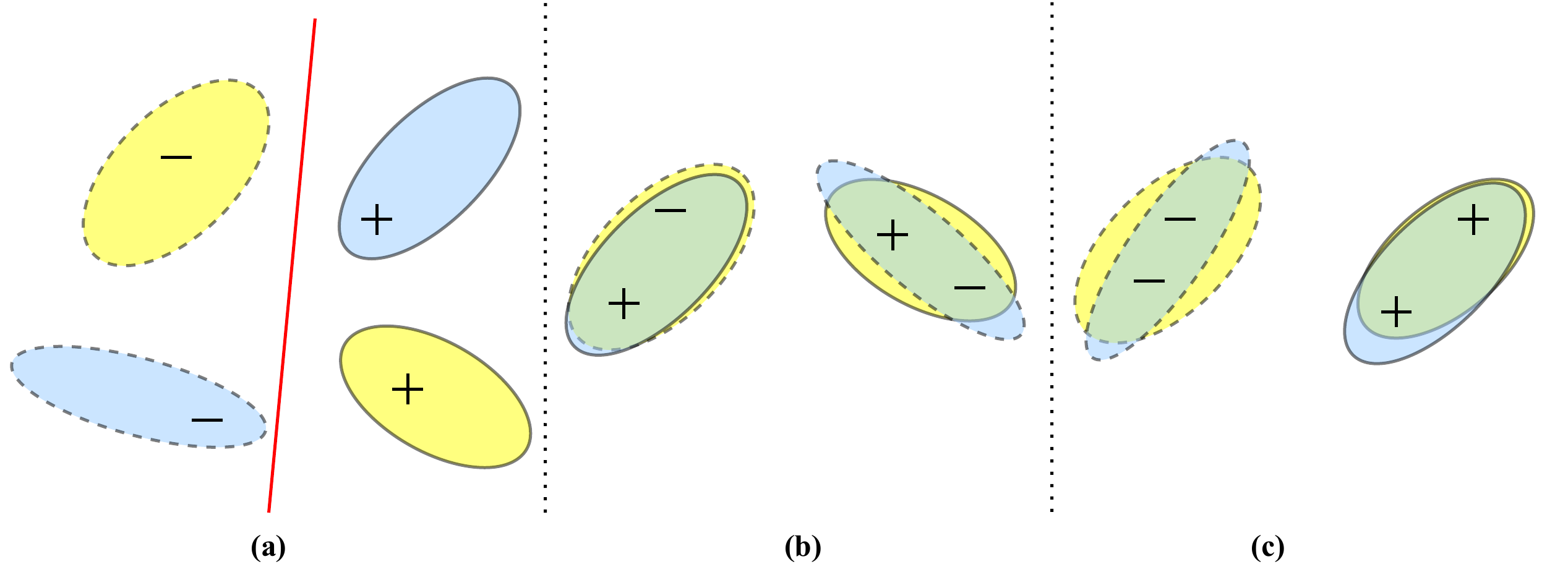}
    \caption{Examples for the domain alignment in MDTC. For simplicity, we present the cases of two domains. The dashed boundary denotes the negative polarity, while the continuous boundary denotes the positive polarity. The blue regions represent the $P_1$ feature distribution and the yellow regions represent the $P_2$ feature distribution. (a) $P_1$ and $P_2$ are not aligned while $d_\mathcal{H}$ is zero for a linear separator presented by the red line; (b) When aligning the marginal distributions, misalignment may occur with regard to the category; (c) Correct alignment is conducted between $P_1$ and $P_2$.
	}
    \label{Fig2}
\end{figure*}

\noindent As a consequence, the main goal of MDTC is to minimize $d_{\mathcal{H}}$. However, a small $d_{\mathcal{H}}$ may not guarantee that the multiple feature distributions are correctly aligned. As illustrated in Fig.\ref{Fig2}(a), it can be noted that $d_{\mathcal{H}}$ can be pushed towards 0 without $P_1$, $P_2$, ..., and $P_M$ being aligned correctly. In addition, only aligning the marginal feature distributions may match instances with different labels, resulting in the weak discriminability of the learned features (Fig. \ref{Fig2}(b)).


In order to address these issues, the proposed CRAL method performs class-conditional alignment by learning two classifiers based on two independent latent spaces ($\mathcal{H}_1$ and $\mathcal{H}_2$), and penalizing the disagreement between their predictions. For any $h_1\in\mathcal{H}_1$, $h_2\in\mathcal{H}_2$, to quantify the disagreement between the two classifiers, we exploit the discrepancy between the predictions $h_1(\xvec)$ and $h_2(\xvec)$ on unlabeled data among the $M$ domains. When the disagreement diminishes to zero, it suggests that the domain alignments performed in these two latent spaces are similar and the $M$ feature distributions are correctly aligned. We here use the $\ell_1$ norm to encode the disagreement:

\begin{align}
    \mathcal{L}_d=\sum_{i=1}^M \mathbb{E}_{\xvec_i\sim\mathbb{U}_i}||\mathcal{G}^{i,1}(\xvec_i)-\mathcal{G}^{i,2}(\xvec_i)||_1
\end{align}

To ensure the sufficient diversity of the two shared latent spaces, we adopt a regularizer formulated as:

\begin{align}
    \mathcal{L}_{\vartriangle}=\min(\gamma,||\frac{1}{M}\sum_{i=1}^M \mathbb{E}_{\xvec_i\sim\mathbb{L}_i} (\mathcal{F}_s^1(\xvec_i)-\mathcal{F}_s^2(\xvec_i))||_2^2) 
\end{align}

\noindent where $||\cdot||_2^2$ is the $\ell_2$ norm, the hyperparameter $\gamma$ is positive and controls the maximal disparity between these two shared latent spaces. Maximizing $\mathcal{L}_{\vartriangle}$ is equivalent to pushing the centroids of these two shared latent spaces far apart.

\subsection{Virtual Adversarial Training}

In this paper, we also incorporate virtual adversarial training (VAT) with entropy minimization to impose consistent prediction constraints on the model. In MDTC, entropy minimization is often utilized to control the uncertainty of their predictions. The entropy minimization is formulated as follows:

\begin{align}
    \mathcal{L}_e^b = -\sum_{i=1}^M\mathbb{E}_{\xvec_i\sim\mathbb{U}_i}[\mathcal{G}^{i,b}(\xvec_i)^{\top}\log(\mathcal{G}^{i,b}(\xvec_i))]
\end{align}

\noindent Even though the entropy minimization can help learn additional structures from unlabeled data to enrich the discriminability of the features learned from labeled data, it should be noted that the entropy minimization may result in overfitting to the unlabeled data if the model has infinite capacity \citep{shu2018dirt}. The over-confident predictions on unlabeled data can make the decision boundary lie across the high-density regions of the feature distributions \citep{french2018self}, deteriorating the discriminability of the learned features. Specifically, as unlabeled data have no supervision, the model may abruptly change its prediction in the vicinity of the unlabeled data \citep{verma2019interpolation} (e.g. for any $h\in\mathcal{H}$, $h(u+\epsilon)\neq h(u)$, where small perturbations $\epsilon$ occur to unlabeled data points $u$). To avoid this risk, virtual adversarial training (VAT) is introduced in conjunction with the entropy minimization in our CRAL to smooth the prediction surface around the unlabeled points. The VAT is proposed to search for small perturbations $\epsilon$ that maximize the change of the predictions on unlabeled data, enforcing consistent predictions to the model \citep{miyato2018virtual}. We also use the VAT on labeled data following \citep{shu2018dirt}. The VAT losses on unlabeled data $\mathcal{L}_{uvt}^b$ and labeled data $\mathcal{L}_{lvt}^b$ are formulated as:

\begin{align}
    \mathcal{L}_{uvt}^b = \sum_{i=1}^M\mathbb{E}_{\xvec_i\sim\mathbb{U}_i}[\max_{||r||\leq\epsilon}D_{kl}(\mathcal{G}^{i,b}(\xvec_i)||\mathcal{G}^{i,b}(\xvec_i+r))]
\end{align}

\begin{align}
    \mathcal{L}_{lvt}^b = \sum_{i=1}^M\mathbb{E}_{\xvec_i\sim\mathbb{L}_i}[\max_{||r||\leq\epsilon}D_{kl}(\mathcal{G}^{i,b}(\xvec_i)||\mathcal{G}^{i,b}(\xvec_i+r))]
\end{align}

\noindent Where $D_{kl}(\cdot||\cdot)$ is the Kullback–Leibler divergence \citep{van2014renyi}. 

\subsection{The Final Objective}

The final objective function of the CRAL method is formulated as:

\begin{equation}
    \begin{aligned}
    \min_{\mathcal{F}_s^b,\{\mathcal{F}_d^{i,b}\}, \mathcal{C}^b}&\max_{\mathcal{D}^b} \sum_{b=1}^2 [\mathcal{L}_c^b + \lambda_{Adv} * \mathcal{L}_{Adv}^b  +\lambda_{lvt} * \mathcal{L}_{lvt}^b \\
    &+\lambda_{uvt} * (\mathcal{L}_{e}^b + \mathcal{L}_{uvt}^b)] + \lambda_d * \mathcal{L}_d  
    - \lambda_{\vartriangle}* \mathcal{L}_{\vartriangle}
    \end{aligned}
\end{equation}

\noindent The CRAL model is trained with backpropagation and in an alternating fashion following \citep{goodfellow2014generative}, the detailed training algorithm is presented in the Supplementary Materials.

\subsection{Discussion}

In MDTC, we have three objectives to accomplish. First, we need to minimize the prediction error on the labeled data; Second, we need to minimize the uncertainty of predictions on the unlabeled data; Third, we should conduct feature alignment among the multiple domains to transform the original features to be domain-invariant. In general, the first objective can be simply achieved by minimizing the cross-entropy loss on the labeled data. How to optimize the second and third objectives are the main challenges of MDTC. Prior MDTC approaches often focus on improving the procedure of adversarial training to optimize the adversarial alignment \citep{wu2020dual,wu2021mixup}, few works investigate how to optimize the uncertainty of predictions on unlabeled data to make advance. SSL suggests that unlabeled data can be used to learn additional structures about the input distributions \citep{grandvalet2005semi}. For instance, cluster structures in the distributions could hint at the separation of samples into different labels. This is termed as the cluster assumption: if two samples reside in the same cluster, they are likely to belong to the same class. The cluster assumption encourages the decision boundary to lie in the low-density regions of the distributions. The intuition is simple: If a decision boundary lies in a high-density region, it will cut the cluster into two different classes, enabling samples lying in the same cluster to have two different labels. The above phenomenon is named as the violation of the cluster assumption. By penalizing the violation of the cluster assumption, we can drive the decision boundary away from the high-density regions of the distributions, imposing consistent prediction constraints on the model.

In our approach, we improve the adversarial training by constructing two adversarial training branches, maximizing the diversity of their shared latent spaces while minimizing the disagreement of their predictions on unlabeled data. This improvement can shrink the search space of possible alignments while preserving the correct set of alignments. Moreover, we optimize the uncertainty of predictions on unlabeled data by employing the VAT with entropy minimization. The VAT can impose consistency regularization on the training data by penalizing the violation of the cluster assumption.

\section{Experiments}

In this section, we first illustrate the datasets and baseline methods. Then we show experimental results on two tasks: MDTC and MS-UDA: The former refers to the setting where the test data fall into one of the $M$ domains; The latter refers to the setting where test data comes from an unseen domain. Finally, the ablation study and parameter sensitivity analysis are provided to give more insights into the CRAL approach.

\subsection{Experimental Settings}

\begin{table*}[t]
\caption{\label{font-table} MDTC classification accuracies on the Amazon review dataset.}\smallskip
\label{table_ref2}
\centering
\resizebox{1.5\columnwidth}{!}{
\smallskip\begin{tabular}{ l|  c c c c c c c c c}
\hline
	Domain & CMSC-LS & CMSC-SVM & CMSC-Log & MAN-L2 & MAN-NLL & DACL & CAN & CRAL(Ours)\\
\hline
Books &  82.10 & 82.26 & 81.81 & 82.46 & 82.98 & 83.45 & 83.76 &  $\mathbf{85.26\pm0.13}$ \\
DVD &  82.40 & 83.48 & 83.73 & 83.98 & 84.03 & 85.50 & 84.68 & $\mathbf{85.83\pm0.14}$ \\
Electr.  & 86.12 & 86.76 & 86.67 & 87.22 & 87.06 & 87.40 & 88.34 & $\mathbf{89.32\pm0.09}$ \\
Kit.  &  87.56 & 88.20 & 88.23 & 88.53 & 88.57 & 90.00 & 90.03 & $\mathbf{91.60\pm0.17}$\\
\hline
AVG  &  84.55 & 85.18 & 85.11 & 85.55 & 85.66 & 86.59 & 86.70 & $\mathbf{88.00\pm0.12}$\\
\hline
\end{tabular}}
\end{table*}

\begin{table*}[t]
\caption{\label{font-table} MDTC classification accuracies on the FDU-MTL dataset.}\smallskip
\label{table_ref3}
\centering
\resizebox{1.8\columnwidth}{!}{
\begin{tabular}{ l| c c c c c c c c c c c}
\hline
	Domain & MT-CNN & MT-DNN & ASP-MTL & BERT & MAN-L2 & MAN-NLL & DA-MTL & DACL & GLR-MTL & CAN & CRAL(Ours)\\
\hline
books & 84.5 & 82.2 & 84.0 & 87.0 & 87.6 & 86.8 & 88.5 & 87.5 & 88.3 & 87.8 &  $\mathbf{89.3\pm0.3}$ \\
electronics & 83.2 & 88.3 & 86.8 & 88.3 & 87.4 & 88.8 & 89.0 & 90.3 & 90.3 & $\mathbf{91.6}$ & 89.1$\pm$0.5 \\
dvd & 84.0 & 84.2 & 85.5 & 85.6 & 88.1 & 88.6 & 88.0 & 89.8 & 87.3 & 89.5 & $\mathbf{91.0\pm0.2}$ \\
kitchen & 83.2 & 80.7 & 86.2 & 91.0 & 89.8 & 89.9 & 89.0 & 91.5 & 89.8 & 90.8 & $\mathbf{92.3\pm0.2}$ \\
apparel & 83.7 & 85.0 & 87.0 & 90.0 & 87.6 & 87.6 & 88.8 & 89.5 & 88.2 & 87.0 & $\mathbf{91.6\pm0.4}$\\
camera & 86.0 & 86.2 & 89.2 & 90.0 & 91.4 & 90.7 & 91.8 & 91.5 & 89.5 & 93.5 & $\mathbf{96.3\pm0.2}$ \\
health & 87.2 & 85.7 & 88.2 & 88.3 & 89.8 & 89.4 & 90.3 & $\mathbf{90.5}$ & $\mathbf{90.5}$ & 90.4 & 87.8$\pm$0.4 \\
music & 83.7 & 84.7 & 82.5 & 86.8 & 85.9 & 85.5 & 85.0 & 86.3 & 87.5 & 86.9 & $\mathbf{88.1\pm0.1}$ \\
toys & 89.2 & 87.7 & 88.0 & 91.3 & 90.0 & 90.4 & 89.5 & 91.3 & 89.8 & 90.0 & $\mathbf{91.6\pm0.3}$ \\
video & 81.5 & 85.0 & 84.5 & 88.0 & 89.5 & 89.6 & 89.5 & 88.5 & 90.8 & 88.8 & $\mathbf{92.6\pm0.4}$ \\
baby & 87.7 & 88.0 & 88.2 & 91.5 & 90.0 & 90.2 & 90.5 & 92.0 & $\mathbf{92.3}$ & 92.0 & 90.9$\pm$0.2 \\
magazine & 87.7 & 89.5 & 92.2 & 92.8 & 92.5 & 92.9 & 92.0 & 93.8 & 92.3 & 94.5 & $\mathbf{95.2\pm0.4}$ \\ 
software & 86.5 & 85.7 & 87.2 & 89.3 & 90.4 & 90.9 & 90.8 & 90.5 & $\mathbf{91.8}$ & 90.9 & 87.7$\pm$0.4 \\
sports & 84.0 & 83.2 & 85.7 & 90.8 & 89.0 & 89.0 & 89.8 & 89.3 & 87.8 & 91.2 & $\mathbf{91.3\pm0.3}$ \\
IMDb & 86.2 & 83.2 & 85.5 & 85.8 & 86.6 & 87.0 & 89.8 & 87.3 & 87.5 & 88.5 &$\mathbf{90.8\pm0.3}$\\
MR & 74.5 & 75.5 & 76.7 & 74.0 & 76.1 & 76.7 & 75.5 & 76.0 & 72.7 & 77.1 & $\mathbf{77.3\pm0.5}$\\
\hline
AVG & 84.5 & 84.3 & 86.1 & 88.1 & 88.2 & 88.4 & 88.2 & 89.1 & 88.5 & 89.4 & $\mathbf{90.2\pm0.2}$ \\
\hline
\end{tabular} 
}
\end{table*}

\begin{table*}
\caption{\label{font-table} Unsupervised multi-source domain adaptation results on the Amazon review dataset.}\smallskip
\label{table_ref4}
\centering
\resizebox{1.5\columnwidth}{!}{
\begin{tabular}{ l| c c c c c c c c c}
\hline
Domain & mSDA & DANN & MDAN(H) & MDAN(S) & MAN-L2 & MAN-NLL & DACL & CAN & CRAL(Ours) \\
\hline
Books & 76.98 & 77.89 & 78.45 & 78.63 & 78.45 & 77.78 & 80.22 & 78.91 & $\mathbf{82.49}$\\
DVD & 78.61 & 78.86 & 77.97 & 80.65 & 81.57 & 82.74 & 82.96 & 83.37 & $\mathbf{84.30}$\\
Elec. & 81.98 & 84.91 & 84.83 & 85.34 & 83.37 & 83.75 & 84.90 & 84.76 & $\mathbf{86.82}$\\
Kit. & 84.26 & 86.39 & 85.80 & 86.26 & 85.57 & 86.41 & 86.75 & 86.75 & $\mathbf{89.08}$\\
\hline
AVG & 80.46 & 82.01 & 81.76 & 82.72 & 82.24 & 82.67 & 83.71 & 83.45 &  $\mathbf{85.67}$\\
\hline
\end{tabular}}
\end{table*}

\begin{table}
\caption{\label{font-table} Ablation study on the Amazon review dataset.}\smallskip
\label{table_ref5}
\centering
\resizebox{0.95\columnwidth}{!}{
\begin{tabular}{ l| c c c c c }
\hline
Method & Books & DVD & Electr. & Kit. & AVG\\
\hline
CRAL (full)& 85.26 & 85.83 & 89.32 & 91.60 & 88.00 \\
CRAL w/o $\mathcal{L}_d$ & 83.08 & 84.19 & 87.45 & 88.74 & 85.87 \\
CRAL w/o $\mathcal{L}_{\vartriangle}$ & 84.62 & 85.18 & 89.15 & 91.17 & 87.53 \\
CRAL w/o $\mathcal{L}_{uvt}$ & 84.25 & 85.07 & 88.64 & 90.88 & 87.21 \\
CRAL w/o $\mathcal{L}_{lvt}$ & 84.96 & 85.35 & 88.93 & 91.21 & 87.61 \\
\hline
\end{tabular}}
\end{table}


\begin{figure*}[htbp]
\centering
\subfigure[$\lambda_d$]{
\includegraphics[width=.4\columnwidth]{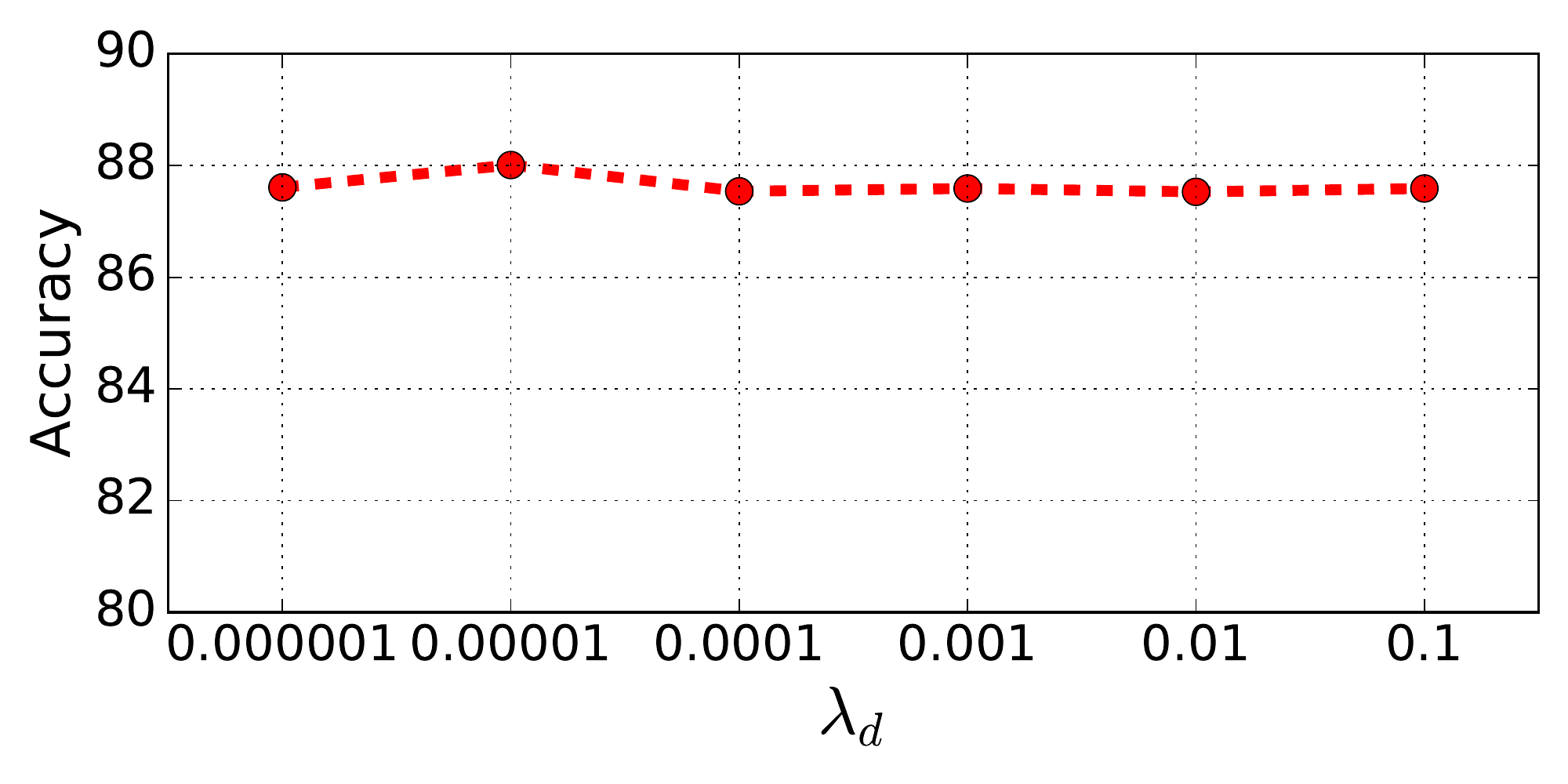}
}
\quad
\subfigure[$\lambda_{\vartriangle}$]{
\includegraphics[width=.4\columnwidth]{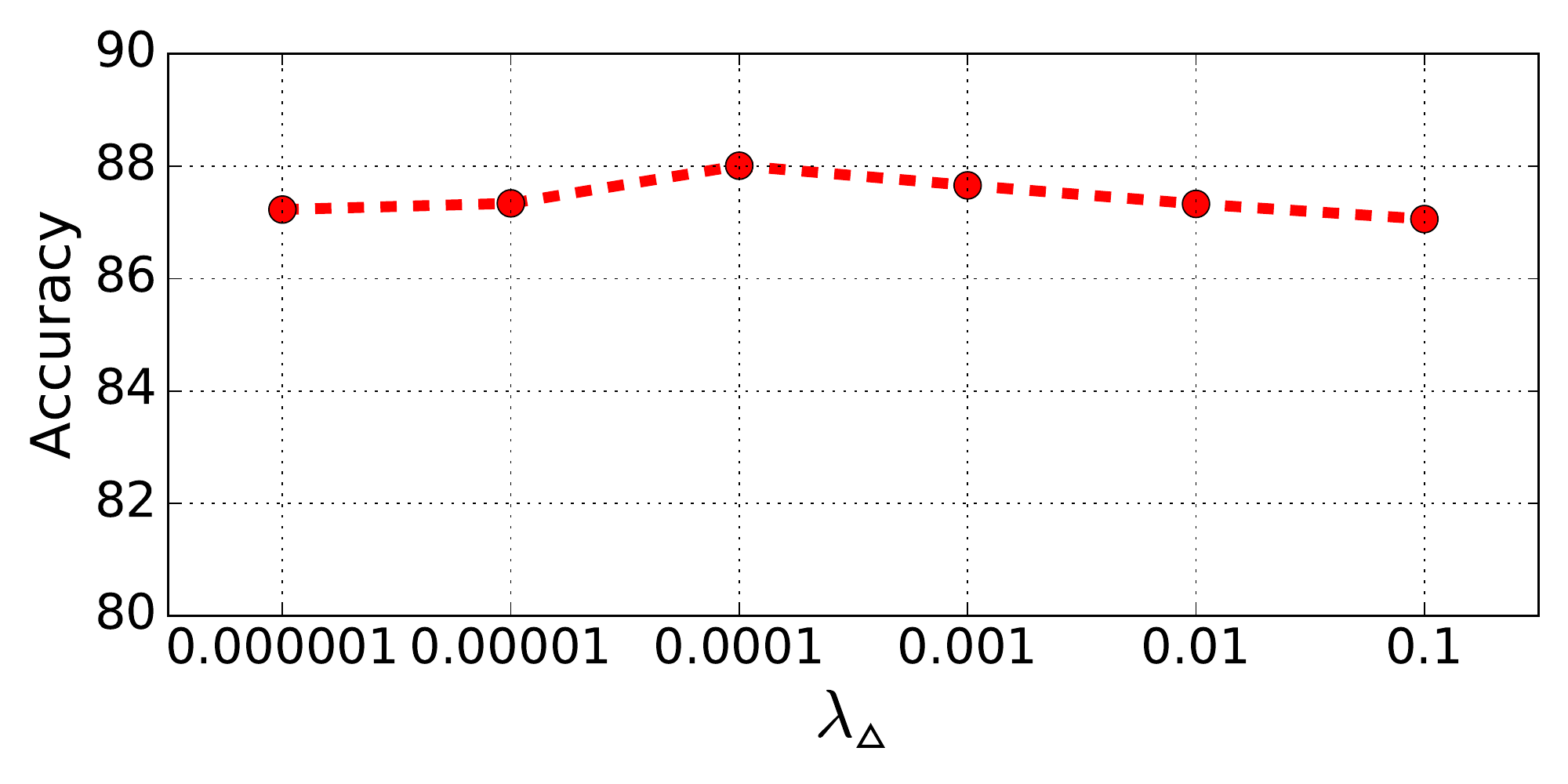}
}
\quad
\subfigure[$\lambda_{uvt}$]{
\includegraphics[width=.4\columnwidth]{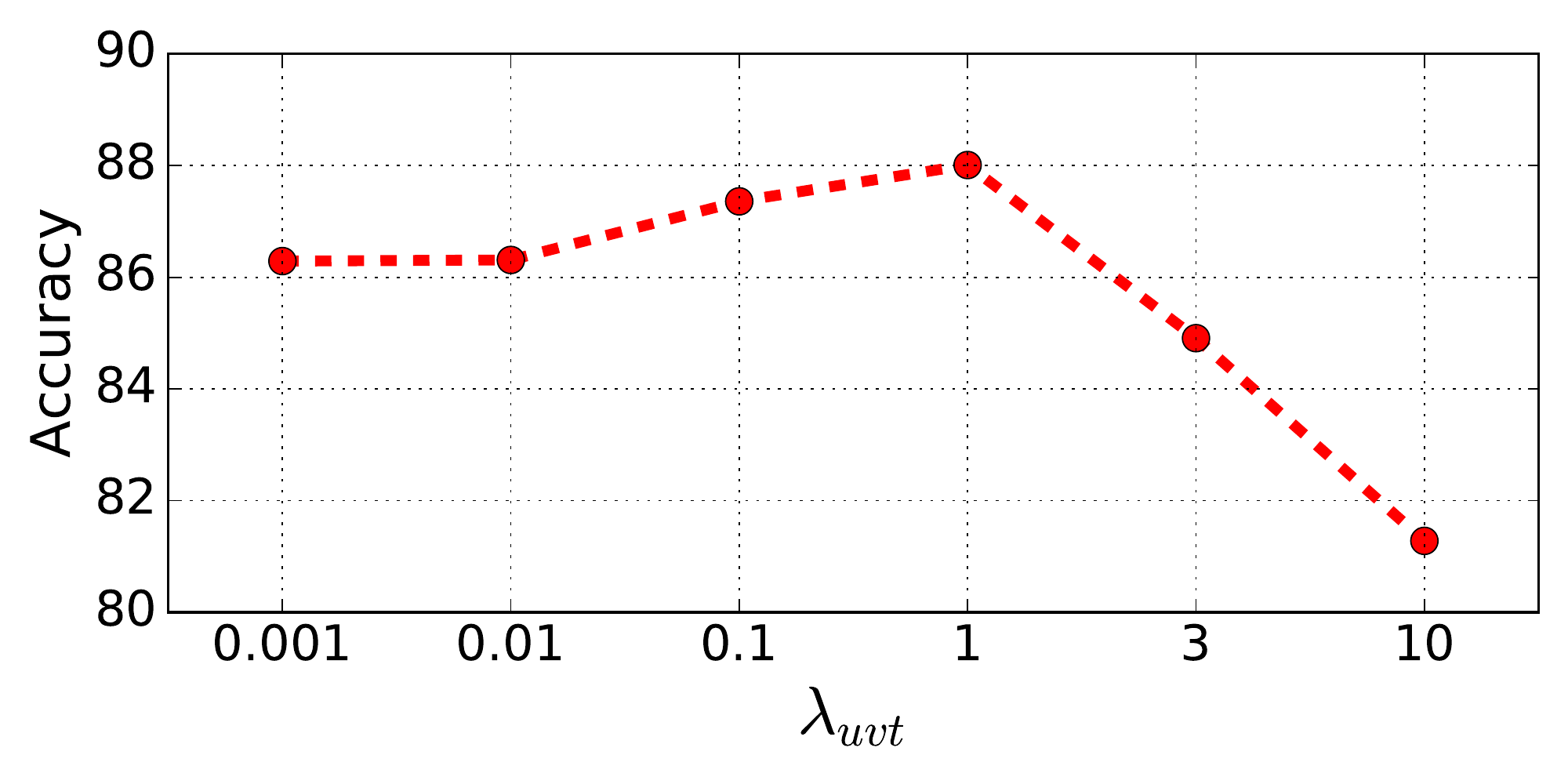}
}
\quad
\subfigure[$\lambda_{lvt}$]{
\includegraphics[width=.4\columnwidth]{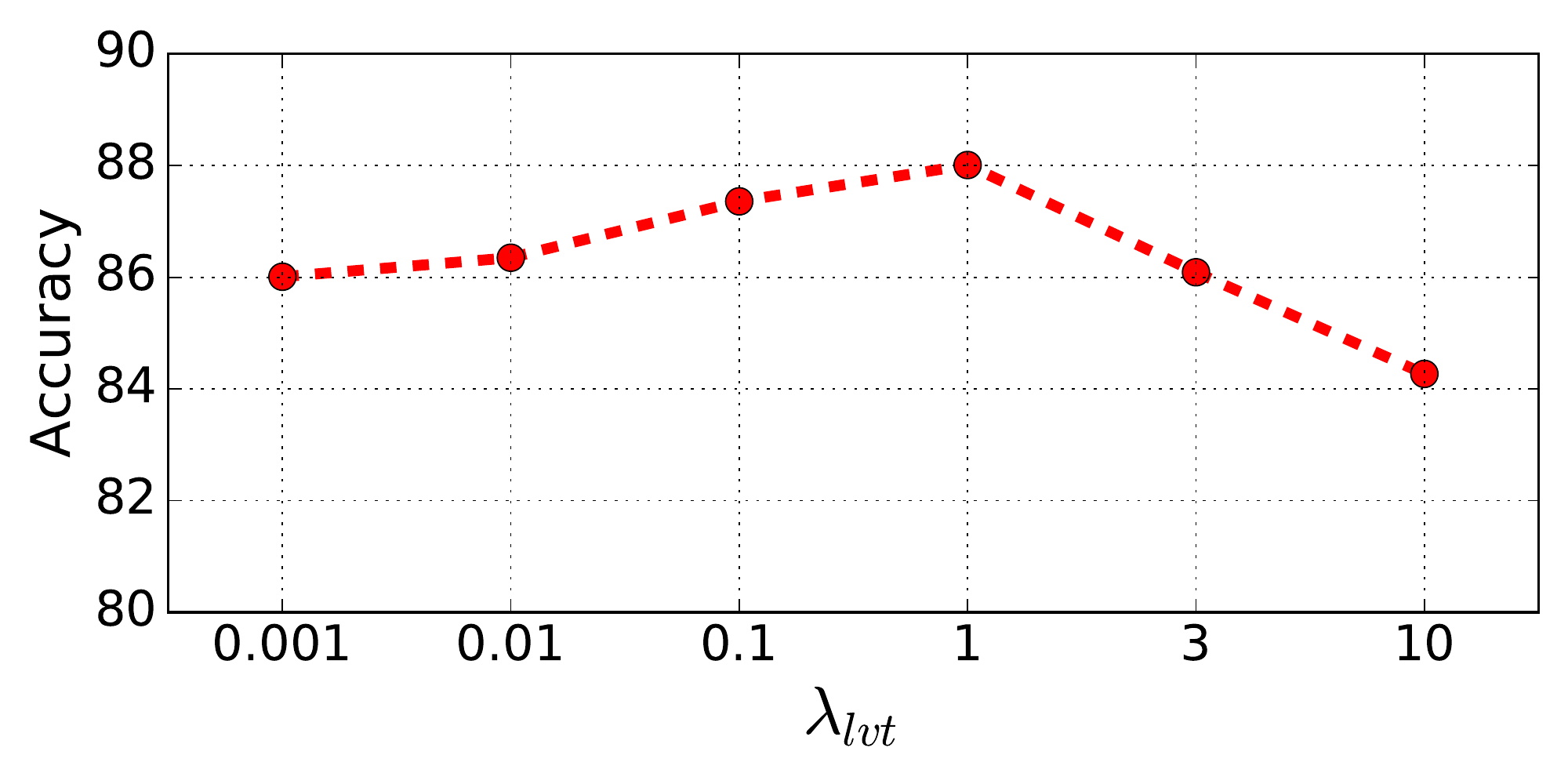}
}
\caption{Parameter sensitivity analysis}
\label{Fig3}
\end{figure*}

\noindent \textbf{Datasets} We conduct experiments on two MDTC benchmarks: the Amazon review dataset \citep{blitzer2007biographies} and the FDU-MTL dataset \citep{liu2017adversarial}. For the Amazon review dataset, there exist 4 domains: books, dvds, electronics, and kitchen. Each domain contains 2000 samples with binary labels: 1000 positive and 1000 negative. All data have been pre-processed into a bag of features (unigrams and bigrams), losing the order information. This prohibits the usage of strong feature extractors (e.g. CNN or RNN). For fair comparisons, we adopt MLPs as the feature extractors and represent each review as a 5000-dimensional vector following \citep{chen2018multinomial}.

As the Amazon review dataset has many limitations, such as the reviews are pre-processed into a bag of features and the lack of word position information. To further validate the effectiveness of the CRAL approach, we also use the FDU-MTL dataset whose data are raw text data, which is in line with the real-world application scenario \citep{liu2017adversarial}. This dataset has 14 product review domains (books, electronics, dvds, kitchen, apparel, camera, health, music, toys, video, baby, magazine, software, and sport) and 2 movie review domains (IMDB and MR). There exist 200 samples in the validation set and 400 samples in the test set for each domain, while the numbers of labeled and unlabeled samples in the training set vary across domains, but are roughly 1400 and 2000, respectively. As demonstrated in \citep{chen2018multinomial}, for experiments on the FDU-MTL dataset, CNN-based feature extractors are much more effective and efficient than LSTM-based feature extractors in learning features. And the most recent baselines (i.e., MAN \citep{chen2018multinomial}, DACL \citep{wu2020dual} and CAN \citep{wu2021conditional}) all adopt CNN-based feature extractors. Therefore, our CRAL method also adopts CNN-based feature extractors in our experiments on the FDU-MTL dataset for fair comparisons.

We take the convenience to cite the experimental results from  \citep{zheng2018same,su2020multi,wu2020dual,wu2021conditional}. The implementation details and the detailed statistics of both datasets are available in the Supplementary Materials.

\noindent \textbf{Comparison Methods} For the MDTC tasks, we compare the CRAL method with a number of state-of-the-art methods: the multi-task convolutional neural network (MT-CNN) \citep{collobert2008unified}, the multi-task deep neural network (MT-DNN) \citep{liu2015representation}, the collaborative multi-domain sentiment classification methods with the least square loss (CMSC-L2), the hinge loss (CMSC-SVM) and the log loss (CMSC-Log) \citep{wu2015collaborative}, the adversarial multi-task learning for text classification (ASP-MTL) \citep{liu2017adversarial}, the multinomial adversarial network with the least square loss (MAN-L2) and the negative log-likelihood loss (MAN-NLL) \citep{chen2018multinomial}, the dynamic attentive sentence encoding method (DA-MTL) \citep{zheng2018same}, the pre-trained BERT-base model which is fine-tuned on each domain (BERT) \citep{devlin2019bert}, the dual adversarial co-learning method (DACL) \citep{wu2020dual}, the global and local shared representation based dual-channel multi-task learning method (GLR-MTL) \citep{su2020multi}, and the conditional adversarial network (CAN) \citep{wu2021conditional}. For MS-UDA experiments, the baselines are listed as follows: the marginalized denoising autoencoder (mSDA) \citep{chen2012marginalized}, the domain adversarial neural network (DANN) \citep{ganin2016domain}, the multi-source domain adaptation network (MDAN) \citep{zhao2017multiple}, the MAN (MAN-L2 and MAN-NLL) \citep{chen2018multinomial}, the DACL \citep{wu2020dual}, and the CAN \citep{wu2021conditional}. As the former two methods are domain-agnostic methods, we combine all training data from the $M$ domains as the training set to train them. 

\subsection{Results}

\noindent \textbf{Multi-Domain Text Classification} We report the classification results of mean $\pm$ standard error over five random trials. The experimental results on the Amazon review dataset and FDU-MTL dataset are reported in Table \ref{table_ref2} and Table \ref{table_ref3}, respectively. From Table \ref{table_ref2}, it can be noted that the CRAL method can not only outperform other baselines on all four domains, but also obtain the best performance in terms of the average classification accuracy. Moreover, it beats the second-best approach CAN by the margin of 1.30\% in terms of the average accuracy.

As shown in Table \ref{table_ref3}, when we conduct experiments on the more challenging FDU-MTL dataset, we can observe that the CRAL method obtains the best classification accuracy on 12 out of 16 domains, and yield the best average classification accuracy, outperforming the second-best baseline CAN by 0.8\%. The results shown in Table \ref{table_ref2} and Table \ref{table_ref3} both demonstrate the effectiveness of our proposed CRAL method in MDTC.

\noindent \textbf{Multi-Source Unsupervised Domain Adaptation} In real application scenarios, it is not uncommon that there exist no labeled data in the target domain. Thus, it is of great significance to evaluate MDTC models in such cases. In the MS-UDA setting, we have multiple source domains with both labeled and unlabeled data and one target domain which only has unlabeled data. The model needs to be trained on the source domains and evaluated on the target domain. When evaluating the CRAL method in the MS-UDA setting, as the target data have no supervision, only the domain-invariant feature vectors are fed into the classifiers, the domain-specific vectors are set to 0s.

We conduct the MS-UDA experiments on the Amazon review dataset, following the same setting as \citep{chen2018multinomial}. For each experiment, three out of the four domains are used as the source domains, and the remaining one is used as the target domain. From Table \ref{table_ref4}, we can see that the proposed CRAL method can outperform other baselines not only on each individual domain, but also in terms of the average accuracy. For the average classification accuracy, our CRAL method outperforms the CAN method by a margin of $2.22\%$, suggesting that our CRAL method has a good capacity for transferring knowledge to unseen domains.

\subsection{Further Analysis}

\noindent \textbf{Ablation Study} In order to verify how each component of the CRAL method can impact the performance, we conduct an ablation study on the Amazon review dataset. In particular, we investigate four variants: (1) CRAL w/o $\mathcal{L}_d$, the variant of CRAL without penalizing on the disagreement of predictions on the unlabeled data; (2) CRAL w/o $\mathcal{L}_{\vartriangle}$, the variant of CRAL without enforcing diversity on two shared feature extractors; (3) CRAL w/o $\mathcal{L}_{uvt}$, the variant of CRAL without the VAT on the unlabeled data; (4) CRAL w/o $\mathcal{L}_{lvt}$, the variant of CRAL without the VAT on the labeled data. The comparison results are presented in Table \ref{table_ref5}. We can see that all four variants induce inferior performance, and the full model produces the best results, validating that all these components contribute to the performance improvement of our model. Moreover, Compared with $\mathcal{L}_{\vartriangle}$, $\mathcal{L}_{uvt}$ and $\mathcal{L}_{lvt}$, $\mathcal{L}_d$ makes the most significant contribution to the system performance by a margin of 2.13\%, illustrating the effectiveness of our co-regularized adversarial learning mechanism. The ablation study on the FDU-MTL dataset presents the same conclusion, its details are available in the Supplementary Materials.

\noindent \textbf{Parameter Sensitivity Analysis} In this section, we explore the sensitivity of our approach to the values of hyperparameters $\lambda_d$, $\lambda_{\vartriangle}$, $\lambda_{uvt}$ and $\lambda_{lvt}$. These hyperparameters are used to trade-off different loss functions. We conduct the parameter sensitivity analysis on the Amazon review dataset. When evaluating one hyperparameter, the others are fixed to their default values (e.g. $\lambda_d$=0.00001, $\lambda_{\vartriangle}$=0.0001, $\lambda_{uvt}$=1, $\lambda_{lvt}$=1). $\lambda_d$ and $\lambda_{\vartriangle}$ are tested in the range \{0.000001, 0.00001, 0.0001, 0.001, 0.01, 0.1\}, and $\lambda_{uvt}$ and $\lambda_{lvt}$ are evaluated in the range \{0.001, 0.01, 0.1, 1, 3, 10\}. The experimental results are shown in Fig. \ref{Fig3}. We report the average classification accuracy.

From Fig. \ref{Fig3}, it can be noted that different values of $\lambda_d$ and $\lambda_{\vartriangle}$ cannot significantly influence the system performance. Consequently, we can set $\lambda_d$ and $\lambda_{\vartriangle}$ as 0.00001 and 0.0001 in all experiments. The trends of performance change for $\lambda_{uvt}$ and $\lambda_{lvt}$ are similar: with the increase of $\lambda_{uvt}$ and $\lambda_{lvt}$, the classification accuracy first increases and reaches the optimum, then it decreases rapidly. This analysis illustrates that proper selections of $\lambda_{uvt}$ and $\lambda_{lvt}$ can effectively improve the performance of our model.

\section{Conclusion} 

In this paper, we propose a co-regularized adversarial learning (CRAL) mechanism for multi-domain text classification. This approach constructs two diverse adversarial training branches and penalizes the disagreement of their predictions on unlabeled data to rule out redundant hypothesis classes while preserving correct alignments. Moreover, it introduces the virtual adversarial training with the entropy minimization to penalize the violation of the cluster assumption, imposing consistency regularization to the model. The experimental results show that our proposed CRAL outperforms the state-of-the-art MDTC methods on two benchmarks. Further investigations demonstrate that our model has a good ability to generalize to unseen domains.

\bibliography{References}


\clearpage
\appendix

\thispagestyle{empty}

\onecolumn \makesupplementtitle

\section{Training Procedure}

The training algorithm of CRAL which uses mini-batch stochastic gradient descent (SGD) is presented in Algorithm 1. In each iteration, the input data should be fed into the two branches to train the model. The CRAL mechanism boosts the system performance by punishing the disagreement of predictions induced from the two diverse adversarial training streams and utilizing the VAT with entropy minimization to enforce the consistency regularization to the model. $\gamma$ is a hyperparameter that controls the diversity extent of two shared latent spaces. $\lambda_{Adv}$, $\lambda_d$, $\lambda_{\vartriangle}$, $\lambda_{uvt}$ and $\lambda_{lvt}$ are hyperparameters that balance different loss functions. In our experiments, we set $\gamma$, $\lambda_{Adv}$, $\lambda_d$ and $\lambda_\vartriangle$ as 10, 1, 0.00001 and 0.0001. According to our parameter sensitivity analysis, the selections of $\lambda_{uvt}$ and $\lambda_{lvt}$ can significantly influence the performance of our model. Thus, these two hyperparameters need to be tuned for each benchmark.

\begin{algorithm}
\caption{SGD training algorithm}\label{trainingalg}
\begin{algorithmic}[1]
\STATE{\bf Input:} labeled data $\mathbb{L}_i$ and unlabeled data $\mathbb{U}_i$ in $M$ domains; hyperparameters: $\gamma$, $\lambda_{Adv}$, $\lambda_{d}$, $\lambda_{\vartriangle}$, $\lambda_{uvt}$, $\lambda_{lvt}$
\FOR{number of training iterations}
	\STATE Sample labeled mini-batches from the multiple domains $B^\ell=\{B^\ell_1,\cdots, B^\ell_M\}$.
	\STATE Sample unlabeled mini-batches from the multiple domains $B^u=\{B^u_1,\cdots, B^u_M\}$.
	\STATE Calculate $l_D =\sum_{b=1}^2\lambda_{Adv}\mathcal{L}_{Adv}^b$ on $B^\ell$ and $B^u$;\\
	Update $\mathcal{D}^1$ and $\mathcal{D}^2$ by ascending along gradients $\nabla l_D$.\\[.2ex] 
	\STATE Calculate $loss=\sum_{b=1}^2[\mathcal{L}_c^b+\lambda_{Adv}*\mathcal{L}_{Adv}^b +\lambda_{uvt}*(\mathcal{L}_e^b+\mathcal{L}_{uvt}^b)+\lambda_{lvt}*\mathcal{L}_{lvt}^b]+ \lambda_d*\mathcal{L}_d-\lambda_{\vartriangle}*\mathcal{L}_{\vartriangle}$ on $B^\ell$ and $B^u$;\\
	Update $\mathcal{F}_s^1$, $\mathcal{F}_s^2$, $\{\mathcal{F}_d^{i,1}\}$, $\{\mathcal{F}_d^{i,2}\}$, $\mathcal{C}^1$, $\mathcal{C}^2$ by descending along gradients $\nabla loss$.\\[.2ex]
\ENDFOR
\end{algorithmic}
\end{algorithm}

\section{Dataset}

The experiments are conducted on two benchmarks: The Amazon review dataset \citep{blitzer2007biographies} and the FDU-MTL dataset \citep{liu2017adversarial}. The data in the Amazon review dataset has been pre-processed into a bag of features (unigrams and bigrams), losing all order information, while the data in the FDU-MTL dataset are raw text data only being processed by the Stanford Tokenizer \citep{manning2014stanford}. The Amazon review dataset contains 4 domains: books, dvds, electronics, and kitchen. All four domains are product reviews. The FDU-MTL dataset contains 16 domains: books, electronics, dvds, kitchen, apparel, camera, health, music, toys, video, baby, magazine, software, sport, IMDB, and MR. The first 14 domains are product reviews while the last two are movie reviews. The details of these two datasets are presented in Table \ref{table_ref6} and \ref{table_ref7}, respectively.

\section{Implementation Details}

We follow the standard experimental settings for MDTC \citep{chen2018multinomial}, adopt the same network architectures as in \citep{chen2018multinomial,wu2020dual}, and ensure that all baselines adopt the standard partitions of the datasets. All experiments are implemented by using Pytorch. The CRAL has six parameters: $\gamma$, $\lambda_{Adv}$, $\lambda_d$, $\lambda_{\vartriangle}$, $\lambda_{uvt}$ and $\lambda_{lvt}$. In the experiments, we fix $\gamma=10$, $\lambda_{Adv}=1$, $\lambda_d=0.00001$, $\lambda_{\vartriangle}=0.0001$, $\lambda_{uvt}$ and $\lambda_{lvt}$ are selected in the range $\{0.001, 0.01, 0.1, 1, 3, 10\}$.

The adam optimizer \citep{kingma2014adam}, with the learning rate 0.0001, is used for training in our experiments. The batch size is 8. The dimension of the shared feature representation is 128 while 64 for the domain-specific one. The dropout rate for each component is 0.4. The classifiers and discriminators are MLPs with one hidden layer of the same size as their input (128 + 64 for classifiers and 128 for discriminators). ReLU is used as the activation function. When evaluating the test data, the average prediction probability of the two classifiers is used to determine the final prediction.

When processing the Amazon review dataset, as we represent its review as a 5000-dimensional vector, we follow \citep{chen2018multinomial} to adopt MLPs as the feature extractors, with an input size of 5000. Each feature extractor consists of two hidden layers, with size 1000 and 500, respectively. In addition, five-fold cross-validation is conducted. We divide the data into five folds per domain: three of the five folds are used as the training set, one is the validation set, and the remaining one is the test set. On one trial, the five-fold average test classification accuracy is reported. For the experiments on the FDU-MTL dataset, we also follow \citep{chen2018multinomial} to use CNN with a single convolutional layer as the feature extractor. It uses different kernel sizes (3, 4, 5), and the number of kernels is 200. The input of the convolutional layer is a 100-dimensional vector, obtained by using word2vec \citep{mikolov2013efficient}, for each word in the input sequence. 

\begin{table}
\caption{\label{font-table} Statistics of the Amazon review dataset}\smallskip
\label{table_ref6}
\centering
\resizebox{0.35\columnwidth}{!}{
\begin{tabular}{ l| c c c }
\hline
Domain & Labeled & Unlabeled & Class.\\
\hline
Books & 2000 & 4465 & 2\\
Electronics & 2000 & 3586 & 2\\
DVD & 2000 & 568 & 2\\
Kitchen & 2000 & 5945 & 2\\
\hline
\end{tabular}}
\end{table}

\begin{table}
\caption{\label{font-table} Statistics of the FDU-MTL dataset}\smallskip
\label{table_ref7}
\centering
\resizebox{0.50\columnwidth}{!}{
\begin{tabular}{ l| c c c c c c c c c}
\hline
Domain & Train & Dev. & Test & Unlabeled & Avg. L & Vocab. & Class.\\
\hline
Books & 1400 & 200 & 400 & 2000 & 159 & 62K & 2\\
Electronics & 1398 & 200 & 400 & 2000 & 101 & 30K & 2\\
DVD & 1400 & 200 & 400 & 2000 & 173 & 69K & 2\\
Kitchen & 1400 & 200 & 400 & 2000 & 89 & 28K & 2\\
Apparel & 1400 & 200 & 400 & 2000 & 57 & 21K & 2\\
Camera & 1397 & 200 & 400 & 2000 & 130 & 26K & 2\\
Health & 1400 & 200 & 400 & 2000 & 81 & 26K & 2\\
Music & 1400 & 200 & 400 & 2000 & 136 & 60K & 2\\
Toys & 1400 & 200 & 400 & 2000 & 90 & 28K & 2\\
Video & 1400 & 200 & 400 & 2000 & 156 & 57K & 2\\
Baby & 1300 & 200 & 400 & 2000 & 104 & 26K & 2\\
Magazine & 1370 & 200 & 400 & 2000 & 117 & 30K & 2\\
Software & 1315 & 200 & 400 & 475 & 129 & 26K & 2\\
Sports & 1400 & 200 & 400 & 2000 & 94 & 30K & 2\\
IMDB & 1400 & 200 & 400 & 2000 & 269 & 44K & 2\\
MR & 1400 & 200 & 400 & 2000 & 21 & 12K & 2\\
\hline
\end{tabular}}
\end{table}

\begin{table}[t]
\caption{Ablation Study on the FDU-MTL dataset.}\smallskip
\label{table_ref8}
\centering
\resizebox{0.75\columnwidth}{!}{
\begin{tabular}{ l| c c c c c }
\hline
	Domain & CRAL(full) & CRAL w/o $\mathcal{L}_d$ & CRAL w/o $\mathcal{L}_{\vartriangle}$ & CRAL w/o $\mathcal{L}_{uvt}$ & CRAL w/o $\mathcal{L}_{lvt}$ \\
\hline
books & 89.3 & 87.3 & 87.6 & 88.5 & 87.1 \\
electronics & 89.1 & 89.0 & 90.8 & 89.0 & 89.9  \\
dvd & 91.0 & 89.7 & 89.5 & 90.3 & 90.4  \\
kitchen & 92.3 & 93.0 & 90.7 & 91.4 & 93.5 \\
apparel & 91.6 & 91.5 & 91.0 & 90.9 & 91.5 \\
camera & 96.3 & 93.2 & 93.4 & 91.5 & 93.2 \\
health & 87.8 & 88.7 & 89.5 & 90.3 & 89.0  \\
music & 88.1 & 82.5 & 86.9 & 86.9 & 86.5 \\
toys & 91.6 & 89.8 & 90.6 & 87.4 & 88.7 \\
video & 92.6 & 88.7 & 91.3 & 90.0 & 87.2 \\
baby & 90.9 & 89.6 & 89.4 & 90.5 & 90.0 \\
magazine & 95.2 & 94.0 & 94.1 & 93.6 & 93.7 \\ 
software & 87.7 & 89.1 & 90.0 & 87.5 & 90.2 \\
sports & 91.3 & 90.9 & 89.6 & 91.5 & 90.5 \\
IMDb & 90.8 & 87.0 & 88.3 & 89.7 & 88.6 \\
MR & 77.3 & 76.3 & 76.0 & 76.0 & 78.5 \\
\hline
AVG & 90.2 & 88.8 & 89.3 & 89.1 & 89.3  \\
\hline
\end{tabular} 
}
\end{table}

\section{Ablation Study}

Similar to the ablation study carried on the Amazon review dataset, we also investigate four variants in the ablation study on the FDU-MTL dataset: (1) CRAL w/o $\mathcal{L}_d$, the variant of CRAL without penalizing on the disagreement of predictions on the unlabeled data; (2) CRAL w/o $\mathcal{L}_{\vartriangle}$, the variant of CRAL without enforcing diversity on two shared feature extractors; (3) CRAL w/o $\mathcal{L}_{uvt}$, the variant of CRAL without the VAT on the unlabeled data; (4) CRAL w/o $\mathcal{L}_{lvt}$, the variant of CRAL without the VAT on the labeled data. The experimental results are shown in Table \ref{table_ref8}. We observe that all four components contribute to the system improvement. In particular, compared to $\mathcal{L}_{\vartriangle}$, $\mathcal{L}_{uvt}$ and $\mathcal{L}_{lvt}$, $\mathcal{L}_{d}$ makes the most significant contribution.

\end{document}